\documentclass[10pt,twocolumn,letterpaper]{article}

\usepackage{cvpr}              % To produce the CAMERA-READY version
\usepackage{pgfplots}
\pgfplotsset{compat=1.18}

\usepackage{booktabs,colortbl,pifont}
\usepackage[dvipsnames]{xcolor}
\usepackage{algorithm}
\usepackage{algpseudocode}

\definecolor{cvprblue}{rgb}{0.21,0.49,0.74}
\usepackage[pagebackref,breaklinks,colorlinks,allcolors=cvprblue]{hyperref}
\usepackage{multirow}
\usepackage{marvosym}

\newcommand{\figref}[1]{Fig.~\ref{#1}}

\newcommand{\eqnref}[1]{Eqn.~(\ref{#1})}
\newcommand{\secref}[1]{Sec.~\ref{#1}}

\def\paperID{} % *** Enter the Paper ID here
\def\confName{CVPR}
\def\confYear{2026}

\newcommand{\method}{{\text{UFO}}\xspace}
\title{UFO: Unifying Feed-Forward and Optimization-based Methods for Large Driving Scene Modeling}

\author{
~~~~~~Kaiyuan Tan\textsuperscript{\rm \ 1, 2 *}
~~~~~~Yingying Shen\textsuperscript{\rm \ 1 *}
~~~~~~Ziyue Zhu\textsuperscript{\rm \ 1}
~~~~~~Mingfei Tu\textsuperscript{\rm 1}
~~~~~~Haohui Zhu\textsuperscript{\rm \ 1} \\
~~~~~~Bing Wang\textsuperscript{\rm 1}
~~~~~~Guang Chen\textsuperscript{\rm 1}
~~~~~~Hangjun Ye\textsuperscript{\rm 1}~\textsuperscript{\Letter}
~~~~~~Haiyang Sun\textsuperscript{\rm 1 $\dagger$}
\\ \\
~ ~ \textsuperscript{\rm 1}Xiaomi EV
~ ~ \textsuperscript{\rm 2}UIUC
\\[1ex]
\textsuperscript{*}Equal contribution. \textsuperscript{\Letter}Corresponding author. \textsuperscript{$\dagger$}Project leader. 
\\
\small{Project Page: \url{https://wm-research.github.io/UFO}}
}

\begin{document}
% \twocolumn[\maketitle\vspace{0em}\input{sec/_1_teaser.tex}\bigbreak]
\maketitle
\begin{abstract}
    Dynamic driving scene reconstruction is critical for autonomous driving simulation and closed-loop learning. 
    While recent feed-forward methods have shown promise for 3D reconstruction, they struggle with long-range driving sequences due to quadratic complexity in sequence length and challenges in modeling dynamic objects over extended durations. 
    We propose \method, 
    a novel recurrent paradigm that combines the benefits of optimization-based and feed-forward methods for efficient long-range 4D reconstruction. 
    Our approach maintains a 4D scene representation that is iteratively refined as new observations arrive, 
    using a visibility-based filtering mechanism to select informative scene tokens and enable efficient processing of long sequences. 
    For dynamic objects, we introduce an object pose-guided modeling approach that supports accurate long-range motion capture. 
    Experiments on the Waymo Open Dataset demonstrate that our method significantly outperforms both per-scene optimization and existing feed-forward methods across various sequence lengths. 
    Notably, our approach can reconstruct 16-second driving logs within 0.5 second while maintaining superior visual quality and geometric accuracy. 
\end{abstract}    
\section{Introduction}
\label{sec:intro}
\begin{figure}[htb]
    \centering
    \includegraphics[width=\linewidth]{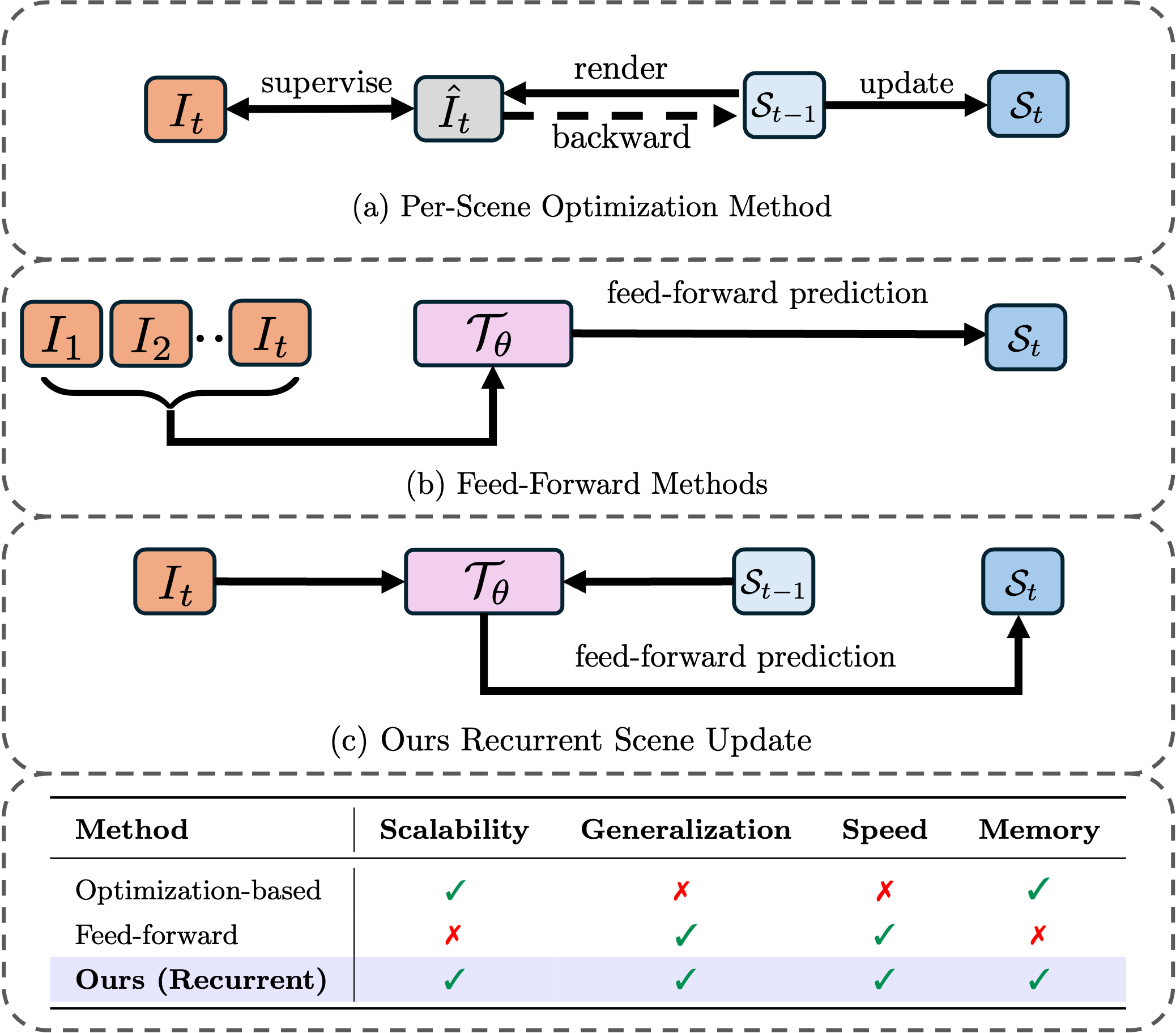}
    \caption{(a) Per-scene optimization methods rely on complex update pipelines to iteratively refine scene representations.  
    (b) Feed-forward methods directly predict 3D representations from image pixels.
    (c) Our \method integrates the strengths of both paradigms by abstracting the render-supervise-update process into a single holistic transformer, enabling efficient long-range 4D reconstruction in a recurrent manner.
    }
    \label{fig:teaser}
    \vspace{-10pt}
\end{figure}

Dynamic driving scene reconstruction is critical for autonomous driving systems, as it enables realistic closed-loop simulation~\cite{hugsim} and reinforcement learning of end-to-end driving algorithms~\cite{gao2025rad}.  
The ability to reconstruct high-fidelity 4D representations of driving scenarios---capturing both spatial geometry and temporal dynamics---is essential for training, and evaluating autonomous systems across diverse, complex, and safety-critical environments.

As illustrated in \figref{fig:teaser}, recent advances in neural scene representation have yielded two primary approaches for 4D reconstruction.
\textbf{Per-scene optimization methods}, based on Neural Radiance Fields (NeRF) \cite{emernerf} or 3D Gaussian Splatting (3DGS) \cite{omnire,streetgs,s3gaussian,pvg}, can achieve high-quality reconstructions by iteratively optimizing scene parameters using photometric losses.
However, these methods suffer from prohibitive computational costs---often requiring hours of optimization per scene---and lack generalization capability, necessitating re-optimization from scratch for each new driving log.
Further, \textbf{Feed-forward methods} \cite{pixelsplat,depthsplat,mvsplat} address these limitations by learning to predict 3D representations directly from input images in a single forward pass, offering much faster inference speed and better generalization with data-driven priors.
Despite their promising results, adapting feed-forward methods to long-range driving scenes remains fundamentally challenging.  
Driving scenarios span large temporal and spatial ranges, featuring high-speed motion and limited inter-camera overlap.  
Recent works~\cite{evolsplat, storm, ggs} mainly handle short sequences with heavily downsampled images due to the quadratic complexity of transformer architectures with respect to spatial resolution and temporal length, making long-sequence processing prohibitively expensive.  
Moreover, modeling dynamic objects over extended durations is difficult.  
Many approaches rely on restrictive assumptions such as constant velocity~\cite{storm}, which fail to capture complex motions.  
Current feed-forward pipelines also lack mechanisms to refine previously reconstructed geometry as new observations arrive, causing error accumulation in long sequences.

To address these challenges, we introduce \method, a novel \textbf{recurrent paradigm for long-range 4D driving scene reconstruction} that synergistically combines the strengths of optimization-based and feed-forward methods.
Our key insight is inspired by the iterative refinement process of 3DGS: rather than reconstructing the entire scene in one shot, we maintain a 4D scene representation composed of scene tokens that are progressively refined as new observations arrive.
At each timestep, our model performs two operations in a learned, feed-forward manner: (1) refinement of existing scene tokens based on new visual evidence, and (2) addition of new scene content not previously captured.
To enable efficient processing of long sequences, we propose a visibility-based filtering mechanism that selects only the most relevant scene tokens for updating at each step, reducing computational complexity from quadratic to near-linear in sequence length.
For dynamic objects, we introduce a object pose-guided modeling approach that leverages 3D bounding boxes from off-the-shelf detectors, combined with learnable per-gaussian lifespans and soft object assignments, enabling accurate long-range motion modeling without restrictive kinematic assumptions.

Extensive experiments on the Waymo Open Dataset demonstrate that \method significantly outperforms both per-scene optimization and state-of-the-art feed-forward methods across various sequence lengths (2s, 8s, 16s).
% Notably, our approach reconstructs 16-second driving logs within 0.5 second while maintaining superior visual quality (PSNR: 27.39) and geometric accuracy (Depth RMSE: 5.10m), with approximately 25\% memory savings compared to existing feed-forward methods on long sequences.
% Our method also exhibits strong zero-shot generalization to unseen sequence lengths and enables effective initialization for per-scene optimization methods.

Our contributions can be summarized as follows:
\begin{enumerate}
\item We introduce a recurrent paradigm for long-range 4D driving scene reconstruction that learns to iteratively refine scene representations in a feed-forward manner, combining the strengths of optimization-based and feed-forward methods.
\item We propose a visibility-based filtering mechanism that achieves near-linear time and memory complexity in sequence length, enabling efficient processing of extended driving sequences. 
\item We present a novel dynamic object modeling approach that leverages object poses with lifespan-aware gaussians, enabling long-range and complex motion capture. 
\end{enumerate}

\section{Related Work}
\label{sec:related}

\paragraph{Feed-Forward Reconstruction}
The field of 3D reconstruction is gradually shifting from multi-stage optimization pipelines towards feed-forward approaches trained on large-scale datasets. Early generalizable NeRFs~\citep{chibane2021stereo,johari2022geonerf,reizenstein2021common,yu2021pixelnerf,suhail2022generalizable,wang2021ibrnet,du2023learning,wang2024distillnerf} typically require extensive point sampling for rendering, resulting in slow inference speed and often limited detail. More recently, feed-forward models based on 3DGS have been proposed~\citep{szymanowicz2024splatter,charatan2024pixelsplat,wewer2024latentsplat,zhang2024gs,szymanowicz2024flash3d,tang2024lgm,xu2024grm}, improving speed, view synthesis quality, and generalization capability. Several works have explored adapting such feed-forward methods to driving-scene reconstruction~\cite{ggs,evolsplat,drivingforward}, but they mainly address static scenes with short, single-camera clips. STORM~\cite{storm} extends feed-forward reconstruction to multi-view dynamic driving scenes, but remains limited to short sequences due to quadratic complexity in sequence length and restrictive kinematic assumptions for object motions. In contrast, our approach recurrently refines a 4D scene representation, enabling linear time complexity over long sequences and models complex, long-range motions through object poses and temporal lifespans. 

Another line of work reconstructs scenes directly from unposed images. DUSt3R~\citep{dust3r} introduces a pointmap representation, while subsequent works extend this direction to dynamic data~\citep{monst3r} and incorporate global attention~\citep{vggt,fast3r} to avoid global alignment post-processing. A related line leverages streaming mechanisms~\citep{spann3r,streamvggt,cut3r,point3r}. Spann3R~\cite{spann3r} and StreamVGGT~\cite{streamvggt} cache historical image features and query them at subsequent timesteps; Point3R~\cite{point3r} employs an explicit point-based memory, whereas CUT3R~\cite{cut3r} uses a fixed-length implicit memory. Although these streaming approaches are conceptually similar to our method, they do not support photorealistic novel view synthesis. Moreover, they are unidirectional: the current frame can only use past frames as reference. In contrast, our method is bidirectional, enabling iterative refinement of the scene at each timestep.

\paragraph{Per-Scene Reconstruction for Driving Scenes}
Built upon recent advances in differentiable volumetric rendering~\cite{nerf,3dgs}, numerous optimization-based approaches have been proposed for reconstructing dynamic driving scenes. These methods can be broadly classified according to how dynamic objects are handled. \textit{Self-supervised} approaches handle dynamic elements implicitly through a unified scene representation, where the separation between static backgrounds and moving entities emerges during optimization. This is typically achieved by modeling scene flow~\cite{suds,emernerf} or deformation over time~\cite{pvg,s3gaussian}. In contrast, explicit decomposition methods model dynamic actor using object-level pose information to enable manipulation and editing. Each dynamic object is modeled in a local coordinate frame given pose information and then transformed into global coordinates for rendering. Methods such as~\cite{s-nerf,mars,unisim,neurad,driving_gaussians,street_gs,hugsim,sgd,omnire,splatad} leverage 3D bounding boxes to isolate dynamic objects, allowing for independent motion modeling and dynamic-object editing. Our method takes inspiration from both paradigms: we utilize bounding boxes for coarse object motion while introducing a lifespan parameterization for finer motion modeling.

\paragraph{Learning to Optimize for NVS}
Several methods leverage data-driven priors for NVS~\citep{flynn2019deepview,chen2024g3r,liu2025quicksplat} but still rely on explicit gradient computation. More closely related to our work, SplatFormer~\citep{splatformer} introduces a refinement network for initialized 3DGS parameters using point transformers, but is limited to object-centric datasets and still suffers from long optimization time. ReSplat~\cite{resplat} proposes a recurrent model that learns to optimize 3D Gaussians in a gradient-free manner using rendering error as feedback; however, it still requires explicit rendering at each iteration and is designed for static scenes. 
In contrast, our method operates in the scene-token space and abstracts away the explicit rendering process, enabling efficient long-range 4D reconstruction in dynamic driving scenes.

\section{Method}
\label{sec:method}

\begin{figure*}[t]
    \centering
    \includegraphics[width=\linewidth]{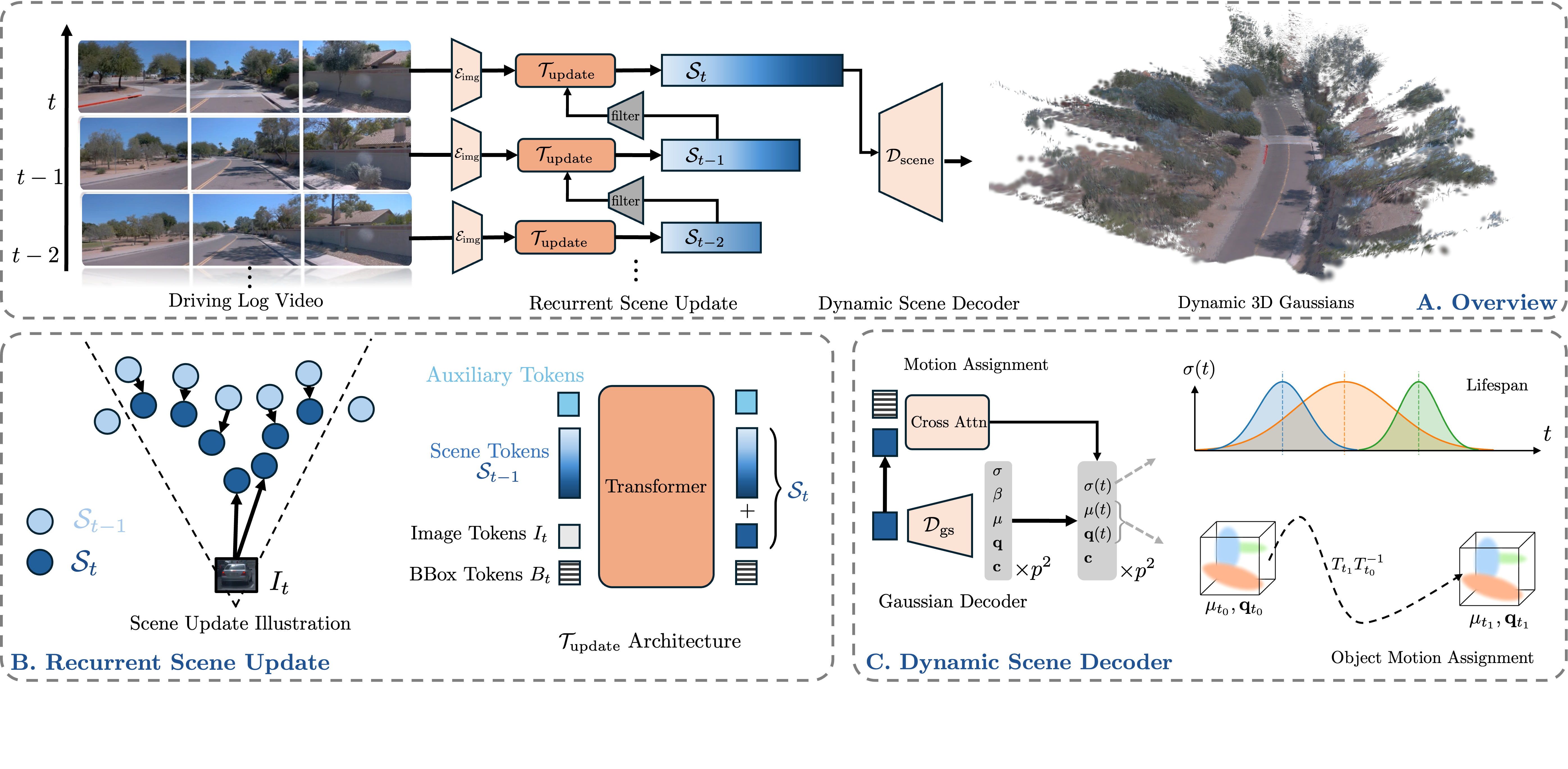}
    \vspace{-40pt}
    \caption{\textbf{Overview of our proposed framework.} Given a long sequence of multi-view images, we reconstruct the 4D scene in a recurrent manner. (A) At each time step, we update the scene representation by refining previous scene tokens based on the new observation and adding new information from the current frame. (B) To efficiently handle long sequences, we employ a visibility-based filtering mechanism to select relevant scene tokens for updating. A unified transformer model learns to update the scene in a feed-forward manner. (C) Dynamic objects are modeled using 3D bounding boxes and per-Gaussian lifespans, enabling complex motion modeling over time.}
    \label{fig:pipeline}
\end{figure*}

\paragraph{Problem Formulation.}
Given a sequence of RGB images $\{I_t\}_{t=1}^T$ captured from a moving camera along with their corresponding camera poses $\{P_t\}_{t=1}^T$, our goal is to efficiently reconstruct a 4D scene representation $\mathcal{S}$ that supports high-quality novel view synthesis (NVS) and accurate depth estimation.

\paragraph{Overview.}
We propose \method, a recurrent paradigm that unifies the strengths of optimization-based and feed-forward methods for long-range 4D reconstruction. As illustrated in \figref{fig:pipeline}, our approach is built upon three key components:

\noindent\textbf{(1) Token-based scene representation} (\secref{sec:scene_rep}): We represent the 4D scene as a compact set of \textit{scene tokens} that are learned from multi-view images through a transformer-based encoder. Each token encodes local geometry, appearance, and motion information, and can be decoded into 3D Gaussians for rendering.

\noindent\textbf{(2) Recurrent scene update} (\secref{sec:iterative_update}): At each time step, when a new frame arrives, we perform two operations in a learned, feed-forward manner: \textit{refining} existing scene tokens based on new visual evidence and \textit{adding} new tokens to capture previously unobserved content. This iterative refinement mimics the optimization loop in 3DGS~\cite{3dgs}, but eliminates explicit rendering or gradient computation, achieving significantly improved efficiency. To enable scalable processing, we introduce a \textit{visibility-based filtering} mechanism that selects only the most relevant tokens for updating, reducing computational complexity from quadratic to near-linear in sequence length.

\noindent\textbf{(3) Dynamic object modeling} (\secref{sec:dynamic_decoder}): We leverage 3D object poses combined with learnable \textit{lifespan parameters} to model complex object motions without restrictive kinematic assumptions.

\subsection{Scene Representation}
\label{sec:scene_rep}

Our scene representation is built upon a token-based paradigm that enables efficient encoding and manipulation of 4D driving scenes. At time step $t$, we maintain a set of scene tokens:
\[
    \mathcal{S}_t = \{\mathbf{s}_t^i\}_{i=1}^{N_t},
\]
where each token $\mathbf{s}_t^i \in \mathbb{R}^{3 + D}$ consists of a 3D location $(x_t^i, y_t^i, z_t^i)$ in world coordinates and a $D$-dimensional feature vector $f_t^i$ ($D = 768$) encoding appearance, geometry, and motion information.

\paragraph{Network Inputs}

% \paragraph{Token Prediction from Images.}
Scene tokens are updated recurrently using a sequence of images and object bounding boxes, along with auxiliary tokens that encodes global information. To obtain an input \textit{(1) image token $I_t$}, we partition the image into patches following ViT~\cite{vit} and construct image tokens by concatenating per-pixel RGB values with Plücker ray embeddings~\cite{plucker} that encode 3D ray geometry. After linear projection and augmentation with time and positional encodings, we obtain image tokens $\mathcal{I}_t$. \textit{(2) Bounding box tokens} $B_t \in \mathbb{R}^{N_{\text{obj}} \times D}$ encode tracked 3D boxes via Fourier embeddings of centers and corners, enabling object-level motion reasoning.
\textit{(3) Auxiliary tokens} are a set of learnable parameters that are used to represent the sky area and per-camera transformations. These tokens are shared across all recurrent steps to aggregate global information. During rendering, sky tokens $\theta_\text{sky}$ are decoded into MLP parameters for sky rendering~\cite{storm}: $\mathbf{c}_\text{sky} = \text{MLP}_\text{sky}(\mathcal{F}(d); \theta_\text{sky})$, and affine tokens $\theta_\text{affine}$ produce per-camera affine transformation for all pixels of a given camera: $\mathbf{c}_\text{affine} = A \cdot \mathbf{c} + b$.
As shown in Figure~\ref{fig:pipeline} part B, at each timestep $t$, these tokens are concatenated into a whole sequence and refined by $\mathcal{T}_\text{update}$ to update the scene representation. 

\subsection{Recurrent Scene Update}
\label{sec:iterative_update}

At the core of our method is a recurrent update mechanism that progressively refines the scene representation as new observations arrive. This design is inspired by the iterative nature of optimization-based methods, but executes entirely in a feed-forward manner. The update module $\mathcal{T}_\text{update}$ is implemented as a 12-layer transformer following the architecture of ViT~\cite{vit}, designed to minimize hand-crafted inductive biases and rely on learned reasoning.

\paragraph{Update Formulation.}
When a new frame $I_t$ with camera pose $P_t$ arrives, we update the scene representation from $\mathcal{S}_{t-1}$ to $\mathcal{S}_t$. Conceptually, a transformer-based update module $\mathcal{T}_\text{update}$ processes the current observation together with existing scene tokens to produce the updated representation:
\begin{align}
    \label{eq:update_naive}
    \mathcal{S}_t = \mathcal{T}_\text{update}(I_t, \mathcal{S}_{t-1}).
\end{align}
This update performs two complementary operations: (1) refining existing tokens from $\mathcal{S}_{t-1}$ based on new visual evidence from $I_t$, and (2) creating new tokens to capture scene content that was not previously observed. Unlike optimization-based methods that require explicit rendering and backpropagation through the rendering process, our model learns to update scene tokens directly from image features, leading to orders of magnitude faster inference.

However, naively applying \eqnref{eq:update_naive} to all tokens becomes computationally prohibitive for long sequences, as the number of tokens $N_t$ grows over time and transformer attention has quadratic complexity. To address this, we introduce a visibility-based filtering mechanism.

\paragraph{Visibility-Based Filtering.}
The key observation is that not all scene tokens are equally relevant for each incoming frame. Tokens that are far from the camera or outside the view frustum provide limited information for the current update. We therefore filter tokens based on their visibility and proximity to the camera.

Specifically, for each incoming frame $I_t$ with pose $P_t$, we first identify all scene tokens that lie within the camera frustum. Among these candidates, we select the closest $K$ tokens based on their distance to the camera center, forming a visible set:
\begin{align}
    \label{eq:visible_set}
    \mathcal{V}_{t-1} = \{\mathbf{s}_{t-1}^i\}_{i=1}^K.
\end{align}
The transformer module then operates only on this filtered subset, producing refined tokens and new tokens:
\begin{align}
    \label{eq:update_visible}
    \Delta \mathcal{S}_t, \mathcal{V}'_{t-1}, = \mathcal{T}_\text{update}(I_t, \mathcal{V}_{t-1}).
\end{align}
The updated scene representation is constructed by replacing the visible tokens with their refined versions, appending the newly created tokens, and keeping all other tokens unchanged:
\begin{align}
    \mathcal{S}_t = (\mathcal{S}_{t-1} \setminus \mathcal{V}_{t-1}) \cup \mathcal{V}'_{t-1} \cup \Delta \mathcal{S}_t.
\end{align}
This design ensures computational resources are focused on the most informative regions while preserving long-term information in background areas. In practice, we set $K = 3600$, which provides an effective balance between reconstruction quality and computational efficiency.

\paragraph{Local Coordinate System.}
To further improve training stability over long sequences, we adopt a local coordinate system at each time step. Before processing, selected scene tokens are transformed from world coordinates into the camera-centric coordinate system of the current frame $I_t$. This transformation centers tokens around the origin and constrains them to a compact spatial range, reducing the dynamic range of token positions and associated ray encodings (e.g., Plücker embeddings~\cite{plucker}).

Empirically, we observe that training in a purely global coordinate system leads to instability on long sequences, likely due to the large variation in token positions and ray representations as the ego-vehicle travels over extended distances. The local coordinate system mitigates this issue and substantially improves convergence for long-range 4D reconstruction.

\subsection{Dynamic Object Modeling}
\label{sec:dynamic_decoder}

Accurate modeling of dynamic objects is crucial for realistic driving scene reconstruction. We address this challenge through a combination of perception-guided motion modeling and temporal lifespan parameterization.

\paragraph{Bounding Box-Guided Motion.}
To enable fine-grained motion modeling and maximal editability of dynamic objects, we leverage 3D bounding boxes obtained from off-the-shelf detection and tracking systems~\cite{street_gs,omnire}. At each time step $t$, we encode the tracked 3D bounding boxes into a set of bounding box tokens $B_t \in \mathbb{R}^{N_{\text{obj}} \times D}$. These tokens are concatenated with scene tokens and image tokens during the update process, allowing the transformer to jointly reason about geometry, appearance, and object-level motion.

During decoding, we assign each scene token a soft probability distribution over all tracked objects via cross-attention. Specifically, the probability of scene token $s$ belonging to object $i$ is computed as:
\begin{align}
    A_{s,i} =
    \frac{\exp\left(q(s)^\top k(B_{t,i})\right)}
         {\sum_{j=1}^K \exp\left(q(s)^\top k(B_{t,j})\right)},
\end{align}
where $q(\cdot)$ and $k(\cdot)$ are learned MLPs that map scene tokens and bounding box tokens to query and key vectors, respectively, and $B_{t,i}$ denotes the embedding of object $i$ at time $t$.

The motion of Gaussians decoded from each scene token is then determined by a weighted average of the corresponding object motions. For rendering at target time $t_1$, the Gaussian parameters (center $\mu$ and rotation $q$) are transformed as:
\begin{align}
    \mu_{t_1} = T_{t_1}T_{t_0}^{-1}\mu_{t_0}, \quad
    q_{t_1}  = T_{t_1}T_{t_0}^{-1}q_{t_0},
\end{align}
where $T_{t_0}$ and $T_{t_1}$ are transformation matrices derived from weighted averages of bounding box transformations:
\begin{align}
    T_{t} = \sum_{i=1}^K A_{s,i} T_{t,i}.
\end{align}
Here, $T_{t,i}$ denotes the transformation matrix of object $i$ at time $t$, and the weights $A_{s,i}$ encode the soft assignment between scene token $s$ and object $i$.

To prevent the model from misusing this soft assignment to fit static background regions, we introduce an object assignment regularization loss that encourages spatial consistency. Specifically, we encourage scene tokens located inside a bounding box to be assigned to the corresponding object through a cross-entropy loss:
\begin{align}
    \mathcal{L}_\text{obj} = -\sum_{s=1}^N \sum_{i=1}^K y_{s,i} \log A_{s,i},
\end{align}
where $y_{s,i} = \begin{cases}1, & \text{if token } s \text{ is inside object } i \\ 0, & \text{otherwise} \end{cases}$.

\paragraph{Temporal Lifespan.}
Inspired by PVG~\cite{pvg}, we further enhance dynamic modeling by introducing a lifespan parameter $\beta$ for each decoded Gaussian. This allows modeling of transient and deformable objects such as pedestrians and cyclists. The opacity of a Gaussian evolves temporally according to:
\begin{align}
    \sigma(t) = \sigma \cdot \exp\left(-\frac{(t - t_0)^2}{2\beta^2}\right),
\end{align}
where $\sigma$ is the base opacity, $t_0$ is the creation time of the Gaussian, and $\beta > 0$ controls its temporal extent. The lifespan parameter is decoded from scene token features and passed through a \texttt{SoftPlus} activation to ensure positivity.

To avoid degenerate solutions where all Gaussians become overly transient, we regularize lifespans to encourage longer temporal persistence:
\begin{align}
    \mathcal{L}_\text{lifespan}
    = \frac{1}{N} \sum_{i=1}^{N} \frac{1}{\beta_i},
\end{align}
where $\beta_i$ is the lifespan of the $i$-th Gaussian. The combination of object pose-guided motion and temporal lifespan enables our model to accurately represent complex dynamic behaviors.

\subsection{Training Objectives}

We train our model end-to-end using a combination of appearance-based and geometric losses. The appearance losses include an L2 photometric loss and a perceptual LPIPS loss~\cite{lpips} to ensure visual quality. For geometric supervision, we employ an L1 depth loss using LiDAR points. Additionally, we apply a sky-mask loss to prevent Gaussians from fitting the sky region. The regularization terms for lifespan and object assignment are also included.

The overall training objective is:
\begin{align}
    \mathcal{L}
    = \mathcal{L}_2
    + \mathcal{L}_\text{LPIPS}
    + \mathcal{L}_\text{depth}
    + \mathcal{L}_\text{lifespan}
    + \mathcal{L}_\text{obj}
    + \mathcal{L}_\text{sky}.
\end{align}

\section{Experiments}
\label{sec:exp}

\paragraph{Implementation Details.}
Our feature decoder is a 12-layer Transformer encoder~\cite{vit} with a patch size of 8 and an embedding dimension of 768. For each frame, we retain up to 32 object boxes, selected by ranking 3D bounding boxes according to the number of LiDAR points inside each box. The box encoder and Gaussian decoder are implemented as 2-layer MLPs. We adopt \texttt{gsplat}~\cite{gsplat} as our Gaussian splatting backend, and use PyTorch's \texttt{flex\_attention} module~\cite{flexattention} to implement attention with custom masks. The total number of trainable parameters is approximately 92M. We train our model on 16 NVIDIA H200 GPUs with a total batch size of 64 for about one day.

\paragraph{Datasets.}
We evaluate novel view synthesis performance on the Waymo Open Dataset (WOD)~\cite{waymo}. Each WOD segment contains roughly 20 seconds driving log data synchronized at 10 Hz. Following \cite{storm}, we use the front, front-left, and front-right cameras in our experiments and sample every 5th frame as context, using the remaining frames for supervision and evaluation. 
\subsection{Visual Quality}
\label{sec:visual_quality}

\begin{figure*}[tb]
    \centering
    \includegraphics[width=0.8\linewidth]{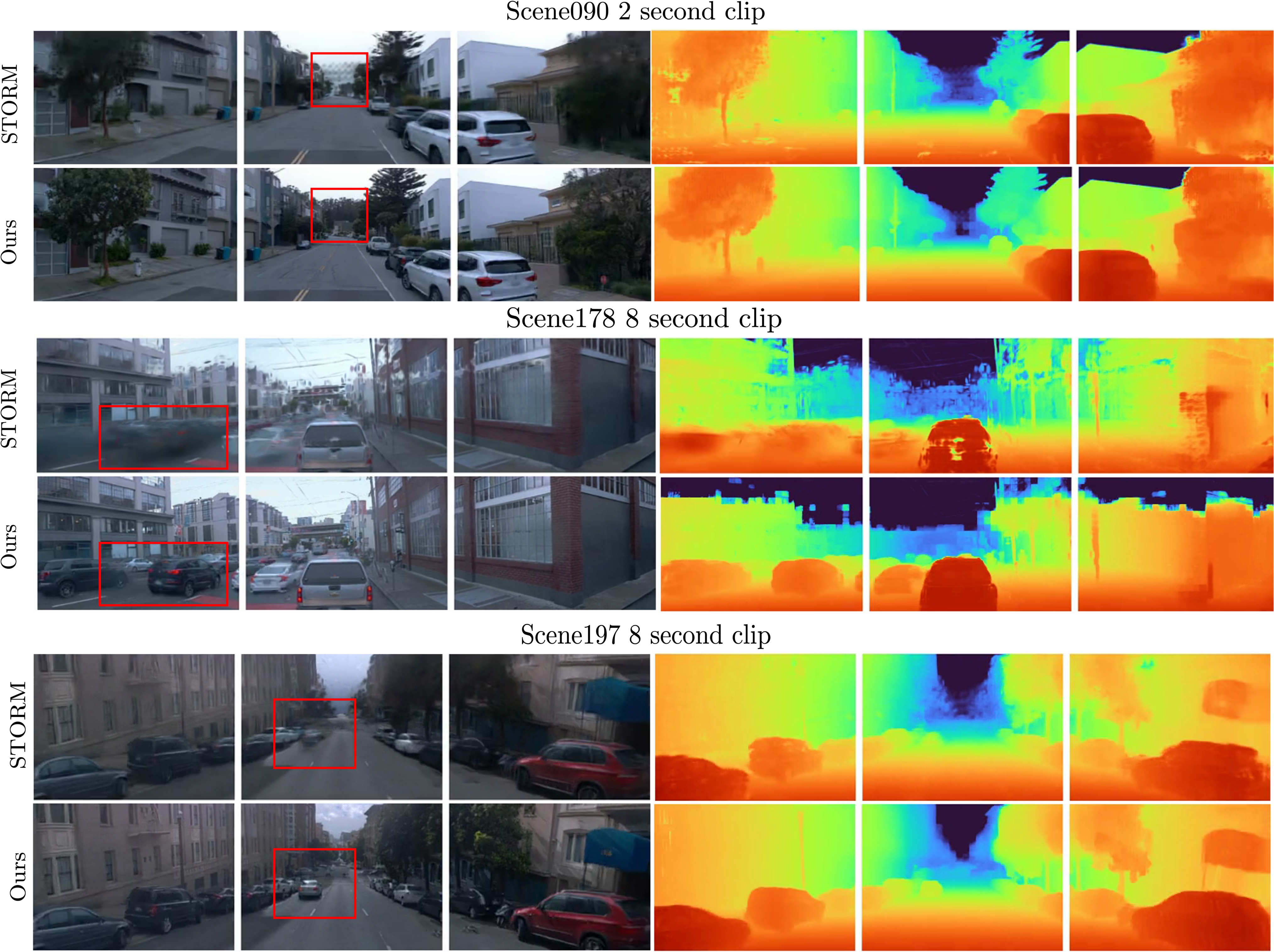}
    \caption{\textbf{Qualitative comparison on novel view synthesis.} }
    \label{fig:qualitative}
\end{figure*}
\paragraph{Baselines.}
We compare against both per-scene optimization methods and generalizable feed-forward methods. The per-scene baselines include 3DGS~\cite{3dgs}, PVG~\cite{pvg}, DeformableGS~\cite{deformablegs}, and Street Gaussians~\cite{streetgs}. The feed-forward baselines include GS-LRM~\cite{gs-lrm} and STORM~\cite{storm}. For optimization-based methods, we use DriveStudio~\cite{omnire} as the implementation codebase. For feed-forward methods, we re-implement GS-LRM and use the official implementations from STORM and adopt model sizes comparable to ours.

Since LiDAR data are not required by the feed-forward approaches, we disable LiDAR-based initialization and depth supervision when training optimization-based methods to ensure a fair comparison. We optimize per-scene methods for 5k, 10k, and 10k iterations for 2s, 8s, and 16s sequences, respectively. We observe that longer optimization does not lead to meaningful improvements. For STORM, we also reproduce the STORM-iterative variant by directly merging the Gaussians predicted from each 2s clip, as described in their paper. For 8s sequences, each feed-forward model is first trained with 2s sequences for 100,000 iterations and fine-tuned with 8s sequences for 50,000 iterations. For 16s sequences, we report zero-shot performance from a model trained only on 8s sequences.\footnote{We could not train STORM or GS-LRM on 16s sequences due to out-of-memory issues.}

\paragraph{Novel View Synthesis.}
Quantitative results are reported in Table~\ref{tab:main}. Across all sequence lengths, our method outperforms both per-scene optimization and feed-forward baselines by a clear margin in terms of PSNR, SSIM, and depth RMSE.

In particular, compared with other feed-forward methods, our approach maintains strong performance as the sequence length increases to 16s, whereas competing feed-forward methods suffer from substantial degradation, even when re-trained on 8s sequences with full global context. Although STORM-iterative preserves fast inference by simply merging short-clip predictions, its visual quality degrades significantly due to the lack of global temporal context. Qualitative comparisons are shown in Figure~\ref{fig:qualitative}. We provide additional qualitative results in the supplementary material. 

\begin{table*}[tb]
    \centering
    \caption{\textbf{Comparison to state-of-the-art methods on the Waymo Open Dataset.} We compare photorealism and geometric metrics against both per-scene optimization methods and feed-forward methods. We report PSNR, SSIM, and Depth RMSE (D-RMSE). 
    }
    \resizebox{\linewidth}{!}{
    \begin{tabular}{lcccccc>{\columncolor{gray!20}}c>{\columncolor{gray!20}}c>{\columncolor{gray!20}}c}
    \toprule
    \multirow{2}{*}{Methods} &
     \multicolumn{3}{c}{2s Sequences}  & \multicolumn{3}{c}{8s Sequences} & \multicolumn{3}{c}{\cellcolor{gray!20}16s Sequences (\textbf{zero-shot})} \\
     & PSNR$\uparrow$ & SSIM$\uparrow$ & D-RMSE$\downarrow$
     & PSNR$\uparrow$ & SSIM$\uparrow$ & D-RMSE$\downarrow$
     & PSNR$\uparrow$ & SSIM$\uparrow$ & D-RMSE$\downarrow$ \\
    \midrule
    % \multicolumn{4}{l}{\textit{Per-Scene Optimization methods}} \\
    3DGS~\citep{3dgs} & 21.07 &  0.578  & 13.52& 19.57& 0.517& 14.42& 17.18& 0.454& 17.01\\
    PVG~\citep{pvg} & 23.81& 0.649& 13.82& 22.90& 0.619& 18.24& 21.79& 0.599& 17.21\\
    Street Gaussians~\citep{streetgs} & 22.96& 0.652& 12.15& 21.69& 0.609& 13.17& \underline{22.67}& \underline{0.675}& 14.88\\
    DeformableGS~\citep{deformablegs} & 23.40& 0.669& 11.55& 21.47& 0.611& 13.12& 19.79& 0.600& 15.82\\
    \midrule 
    % \multicolumn{4}{l}{\textit{Generalizable feed-forward methods}} \\ 
    % LGM~\citep{lgm} & 23.59& 0.691& 8.02& & & & & & \\ 
    GS-LRM~\citep{gslrm} & 25.18& 0.753& 7.94& 21.81& 0.584& \underline{7.37}& 16.98& 0.500& 9.81\\ 
    STORM~\citep{storm} & \underline{26.38}&  \underline{0.794}&  \underline{5.48}& \underline{24.48}& \underline{0.736}& 8.11 & 22.02 & 0.614& \underline{7.91}\\ 
    STORM(iterative)~\citep{storm} & \underline{26.38}&  \underline{0.794}&  \underline{5.48}& 21.25& 0.609& 12.35& 19.88& 0.541& 11.65\\ 
    \textit{Ours}-\method & \textbf{27.26}& \textbf{0.825}& \textbf{5.45}& \textbf{27.39}& \textbf{0.830}& \textbf{5.10}& \textbf{27.04}& \textbf{0.8162}&  \textbf{5.08}\\
    \bottomrule
    \end{tabular}
    }
    \label{tab:main}
\end{table*}

\subsection{Scalability Analysis}

\begin{figure}[tb]
    \centering
    \includegraphics[width=\linewidth]{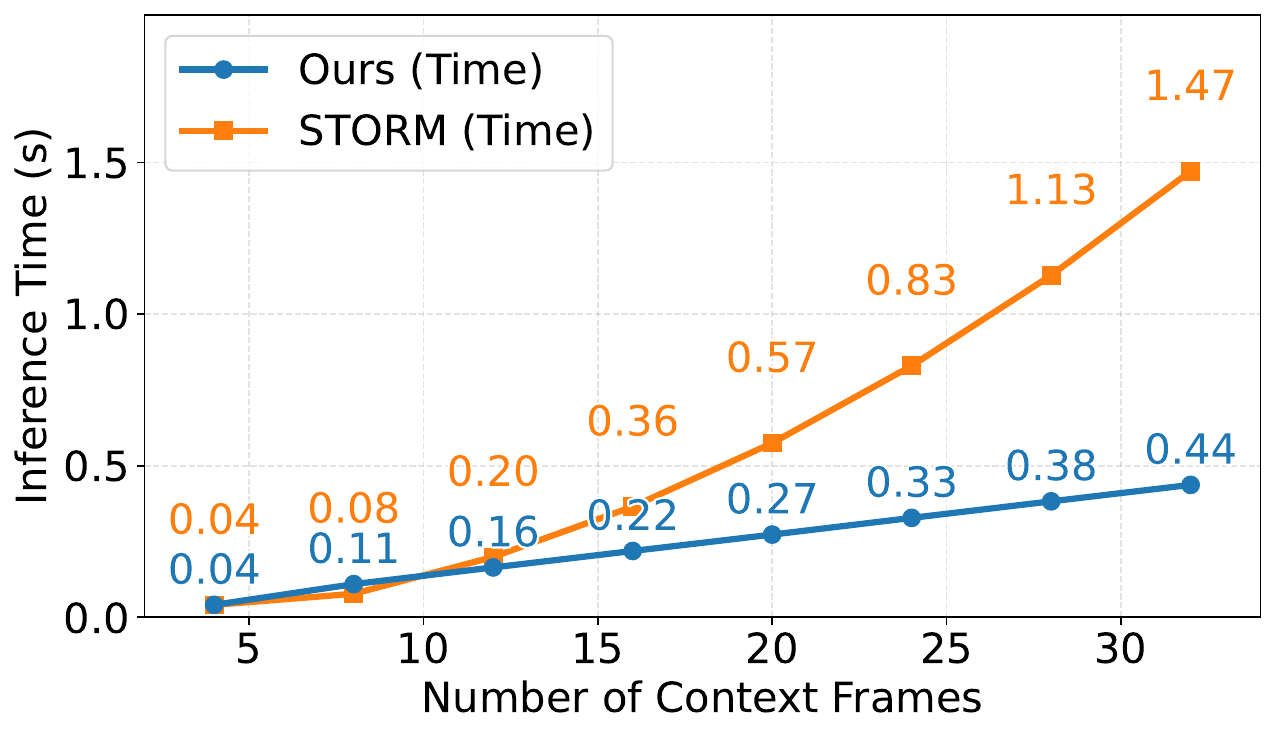}
    \includegraphics[width=\linewidth]{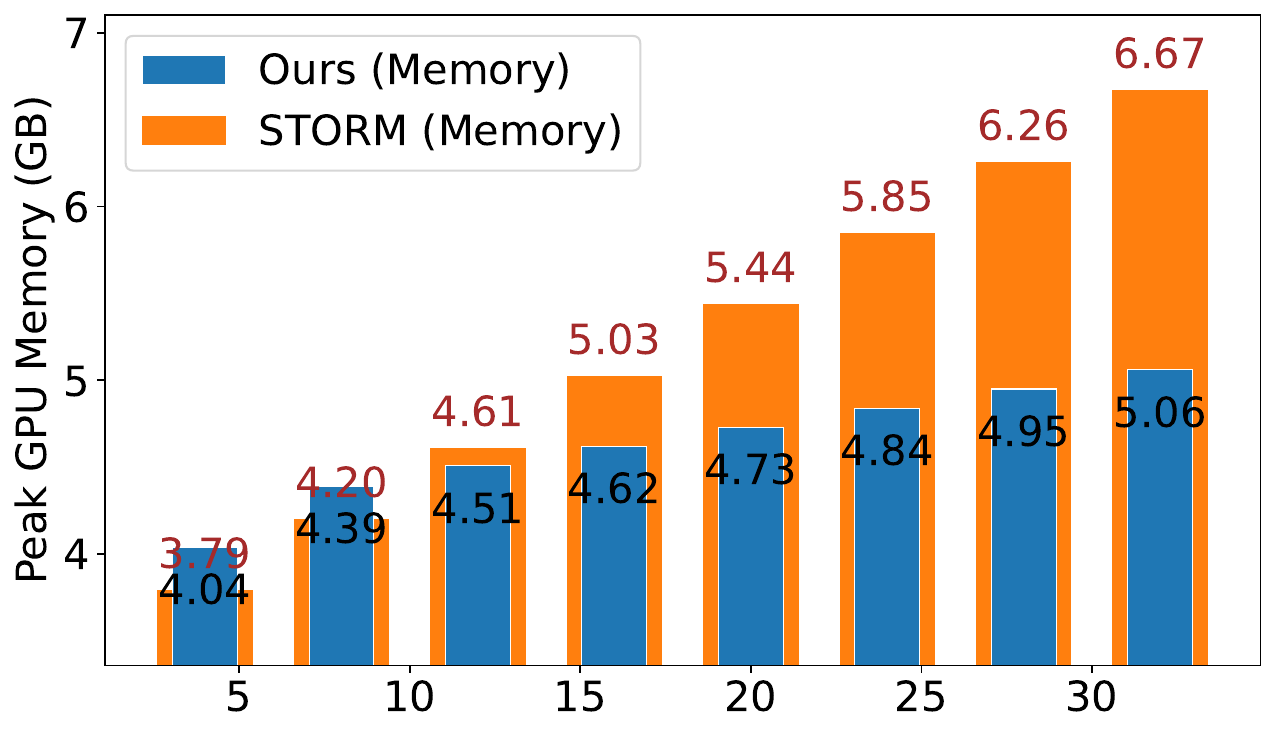}

    \caption{\textbf{Inference Time and Memory Usage Comparison.} We compare inference time and memory usage for different sequence lengths between STORM and our method. Results are reported excluding Gaussian rendering stage.}
    \label{fig:efficiency}
\end{figure}

Computation time and memory consumption are critical for extending the applicability of our method to long sequences. We follow the experimental settings in Section~\ref{sec:visual_quality} and measure inference time and peak GPU memory usage across different sequence lengths. All measurements are conducted on a single NVIDIA H20 GPU with a batch size of 1. The results are reported in Figure~\ref{fig:efficiency}.

Our method exhibits nearly linear time complexity in the input sequence length $n$, while the baseline (STORM) displays quadratic growth. Although both methods show roughly linear memory growth, our method scales more slowly and uses approximately 25\% less memory for 16s sequences. All measurements exclude the Gaussian rendering stage to focus on the behavior of the reconstruction networks.

\subsection{Dynamic Objects Modeling}
Accurately modeling dynamic objects is crucial for downstream applications such as autonomous driving simulation. We evaluate novel view synthesis performance restricted to dynamic objects under different numbers of input frames within a fixed 2-second time span, which defines varying difficulty levels: as the density of input frames increases, the model can access closer temporal context and thus better capture object motion.

\begin{figure}
    \centering
    \includegraphics[width=\linewidth]{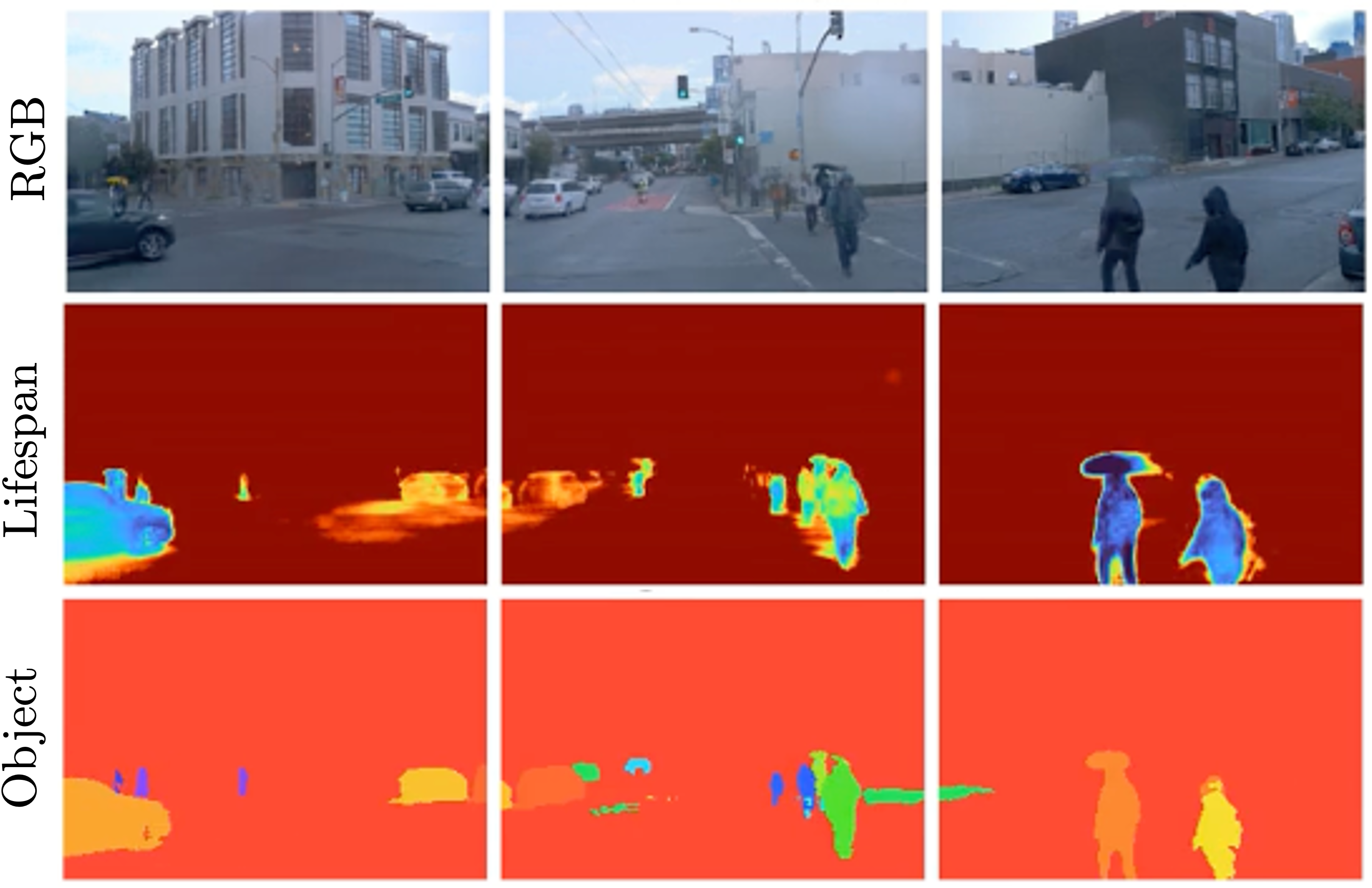}
    \caption{\textbf{Visualization of dynamic object modeling.} Lifespan and motion assignment enable accurate modeling of dynamic objects. Top: rendered RGB at a novel timestep. Middle: rendered lifespan map, where blue indicates transient objects with short lifespan. Bottom: rendered motion-assignment map; different colors indicate different objects. We leverage object poses to transform Gaussians accordingly.}
    \label{fig:dynamic_vis}
\end{figure}

\begin{figure}[tb]
    \centering
    \includegraphics[width=\linewidth]{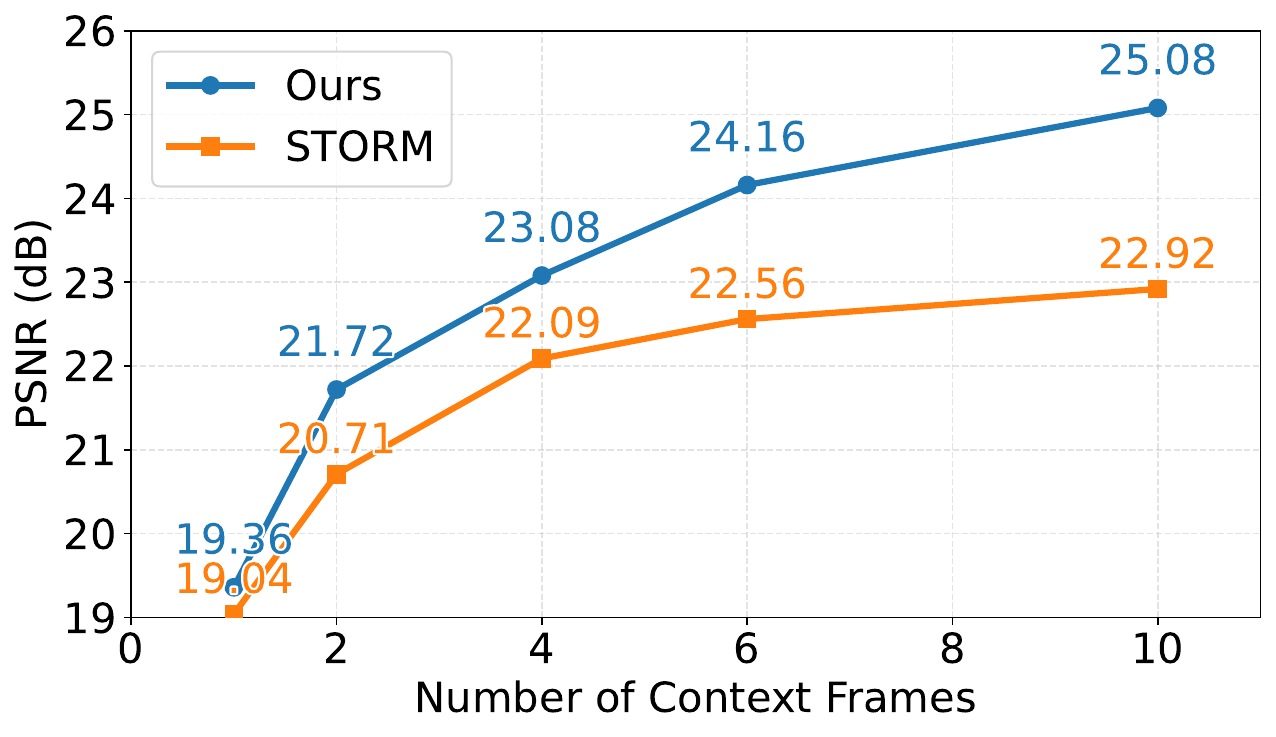}
    \caption{\textbf{Dynamic Object Modeling Performance.} We compare novel view synthesis performance on dynamic objects with different number of input frames within a fixed timespan of 2 seconds.}
    \label{fig:dynamic_objects}
\end{figure}

To isolate the effect of dynamic scene modeling, we disable the recurrent scene update and use a single-step prediction setup. The results are reported in Figure~\ref{fig:dynamic_objects}. Across all input-frame configurations, our method consistently outperforms STORM. This indicates that bounding-box guidance combined with lifespan-aware Gaussian modeling more effectively captures complex object motions than STORM's constant-velocity assumption and velocity-basis parameterization. Qualitative results in Figure~\ref{fig:qualitative} confirm the effectiveness of our approach over assumed constant velocity motion. A demonstration of our dynamic object modeling method is shown in Figure~\ref{fig:dynamic_vis}.

\subsection{Ablation Studies}

We conduct ablation studies to validate the effectiveness of the key components in our method. We report the results in Table~\ref{tab:ablation}. 
In experiment (1) we only provide 2s sequences during training and the model predicts without scene tokens. 
In experiment (2) we remove the scene tokens and each chunk is predicted independently, all chunks are concatenated and rendered for supervision. In experiment (3) we disable the refinement operation and only add new Gaussians at each timestep. In experiment (4) we remove the lifespan prediction and set all Gaussian lifespans to cover the entire sequence. All evaluation are conducted on 8s WOD sequences with 16 uniformly sampled input frames.

\begin{table}[htb]
\centering
\small

\begin{tabular}{lccc} 
\toprule
Method & PSNR$\uparrow$ & SSIM$\uparrow$ & D-RMSE$\downarrow$ \\
\midrule
(1) w/o iterative training & 20.67 & 0.625 & 11.81 \\
(2) w/o scene tokens       & 26.31 & 0.787 &  5.31 \\
(3) w/o refinement         & 27.08 & 0.824 &  5.13 \\
(4) w/o lifespan           & 26.47 & 0.790 &  5.21 \\
(5) w/o bounding box       & 26.88 & 0.818 &  5.12     \\
\midrule
Ours (full)                & \textbf{27.39} & \textbf{0.830} & \textbf{5.10} \\
\bottomrule
\end{tabular}
\caption{Ablation study on key design components. }
\label{tab:ablation}
\end{table}

We observe substantial performance degradation when scene tokens are not refined or are removed entirely, highlighting the importance of our recurrent refinement paradigm. Removing lifespan prediction (variant (4)) also hurts performance, confirming that explicit temporal modeling of Gaussian opacity improves dynamic object reconstruction. Likewise, the performance drop in the \emph{w/o bounding box} variant indicates that explicit bounding-box guidance is important for accurate dynamic object motion.

Interestingly, even without scene tokens in variant (2), the model still benefits noticeably from iterative training across temporal chunks. Empirically, we find that the model learns a prior over the correlation between opacity and time: earlier timesteps tend to produce short-range Gaussians, while later timesteps produce long-range Gaussians. This learned temporal prior helps allocate Gaussians more effectively across the sequence and contributes to overall performance gains, even in the absence of explicit scene-token memory.

\section{Conclusion}
We present \method, a novel recurrent paradigm that unifies the strengths of optimization-based and feed-forward methods for efficient long-range 4D driving scene reconstruction. By maintaining a progressively refined 4D scene representation and employing a visibility-based filtering mechanism, our approach achieves near-linear complexity in sequence length while delivering superior visual quality and geometric accuracy. Extensive experiments on the Waymo Open Dataset demonstrate that \method significantly outperforms existing methods across various sequence lengths, enabling rapid reconstruction of extended driving logs with reduced memory consumption. Our work paves the way for practical application of data-driven reconstruction approaches in autonomous driving simulation and closed-loop learning.

% \newpage
{
    \small
    \bibliographystyle{ieeenat_fullname}
    \bibliography{main}
}

\clearpage
\setcounter{page}{1}
\maketitlesupplementary

% Compact spacing to fit 2 pages
\setlength{\textfloatsep}{8pt plus 1pt minus 1pt}
\setlength{\floatsep}{8pt plus 1pt minus 1pt}
\setlength{\intextsep}{8pt plus 1pt minus 1pt}
\setlength{\abovecaptionskip}{4pt}
\setlength{\belowcaptionskip}{0pt}

\section{Video Results}
We provide video demonstrations of our method's reconstruction quality and dynamic modeling capabilities in the \texttt{video} folder. Each video showcases novel view synthesis results on sequences of varying lengths from the Waymo Open Dataset validation set. The videos include renderings of RGB images, depth maps, as well as visualizations of predicted lifespan maps and object motion assignments to illustrate how our method handles transient phenomena and dynamic objects.

The provided video files follow the naming convention \texttt{[scene\_idx]\_[ctx\_freq]\_[length].mp4}, where: \texttt{scene\_idx} is the index of the scene in the validation set. \texttt{ctx\_freq} is the number of context frames used per second (maximum: 10 fps). \texttt{length} is the total sequence duration in seconds.

\section{Additional Dataset \& Implementation Details}
We train and evaluate our model using the Waymo Open Dataset~\cite{waymo}, following their official train-validation split. We downsample the images to 160$\times$240 and report results on the front, front-left, and front-right cameras, consistent with previous work~\cite{storm}. The model is trained on 16 NVIDIA H200 GPUs with a total batch size of 64 for approximately one day using the AdamW optimizer.

We warm up the model for 5,000 iterations with a linearly increasing learning rate from 0 to $1 \times 10^{-4}$, after which a constant learning rate of $1 \times 10^{-4}$ is applied. Our training objective consists of the following loss components:
\begin{enumerate}
    \item[-] \texttt{RGB MSE Loss} ($\lambda_{\text{RGB}} = 1.0$): Base reconstruction loss.
    \item[-] \texttt{Perceptual Loss} ($\lambda_{\text{LPIPS}} = 0.05$): LPIPS loss, activated after iteration 5000.
    \item[-] \texttt{Depth Loss} ($\lambda_{\text{depth}} = 1.0$): L1 loss on LiDAR depth.
    \item[-] \texttt{Flow Regularization} ($\lambda_{\text{flow}} = 0.005$): MSE on forward flow.
    \item[-] \texttt{Lifespan Regularization} ($\lambda_{\text{lifespan}} = 0.0001$): Encourages persistent Gaussians.
    \item[-] \texttt{Sky Depth Loss} ($\lambda_{\text{sky-depth}} = 0.01$): MSE for sky regions.
    \item[-] \texttt{Sky Opacity Loss} ($\lambda_{\text{sky-opacity}} = 0.1$): L1 loss on sky opacity.
    \item[-] \texttt{Object Assignment Loss} ($\lambda_{\text{obj}} = 1.0$): Cross-entropy for object classification.
\end{enumerate} 

\section{Adversarial Lighting Conditions}
Beyond modeling deformable objects, the lifespan parameter $\beta$ naturally handles transient lighting artifacts such as lens flare, windshield reflections, and sunlight glare. These phenomena are temporally unstable and view-dependent, causing traditional methods to incorrectly bake them into static geometry, resulting in ghosting effects that degrade visual and geometric quality.

\begin{figure}[!h]
    \centering
    \includegraphics[width=\linewidth]{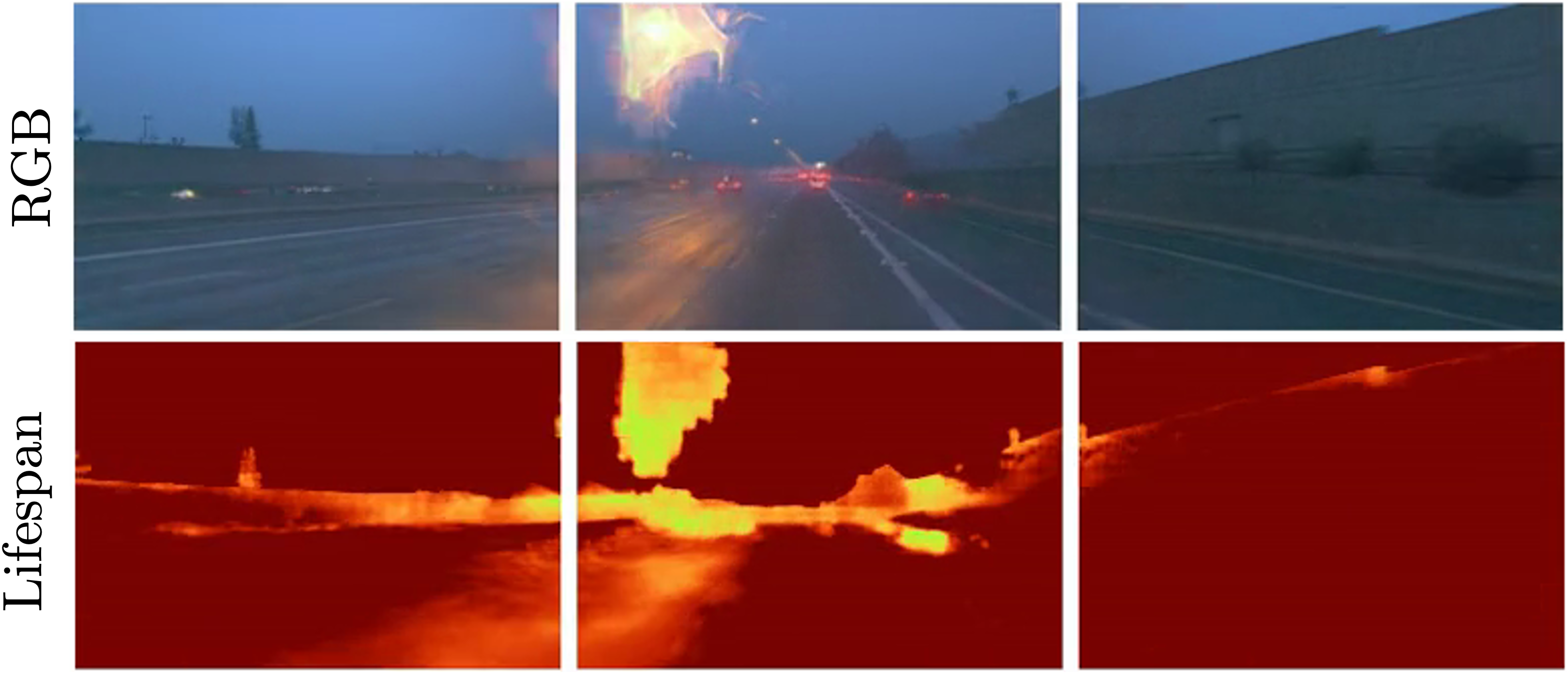}
    \caption{Results under Adverse Lighting Conditions.}
    \label{fig:lifespan}
\end{figure}

Our method addresses this through learned temporal consistency. As shown in Figure~\ref{fig:lifespan}, when the model observes localized bright regions appearing inconsistently across viewpoints and timesteps, it automatically assigns very short lifespans to the corresponding Gaussians, causing them to decay rapidly outside their narrow temporal window. Conversely, genuine scene content with high spatio-temporal consistency receives long lifespans, ensuring stable geometry. This behavior emerges naturally from our training objective without explicit artifact detection or supervision.

\section{Full Qualitative Comparisons}

\begin{figure*}
    \centering
    \includegraphics[width=\linewidth]{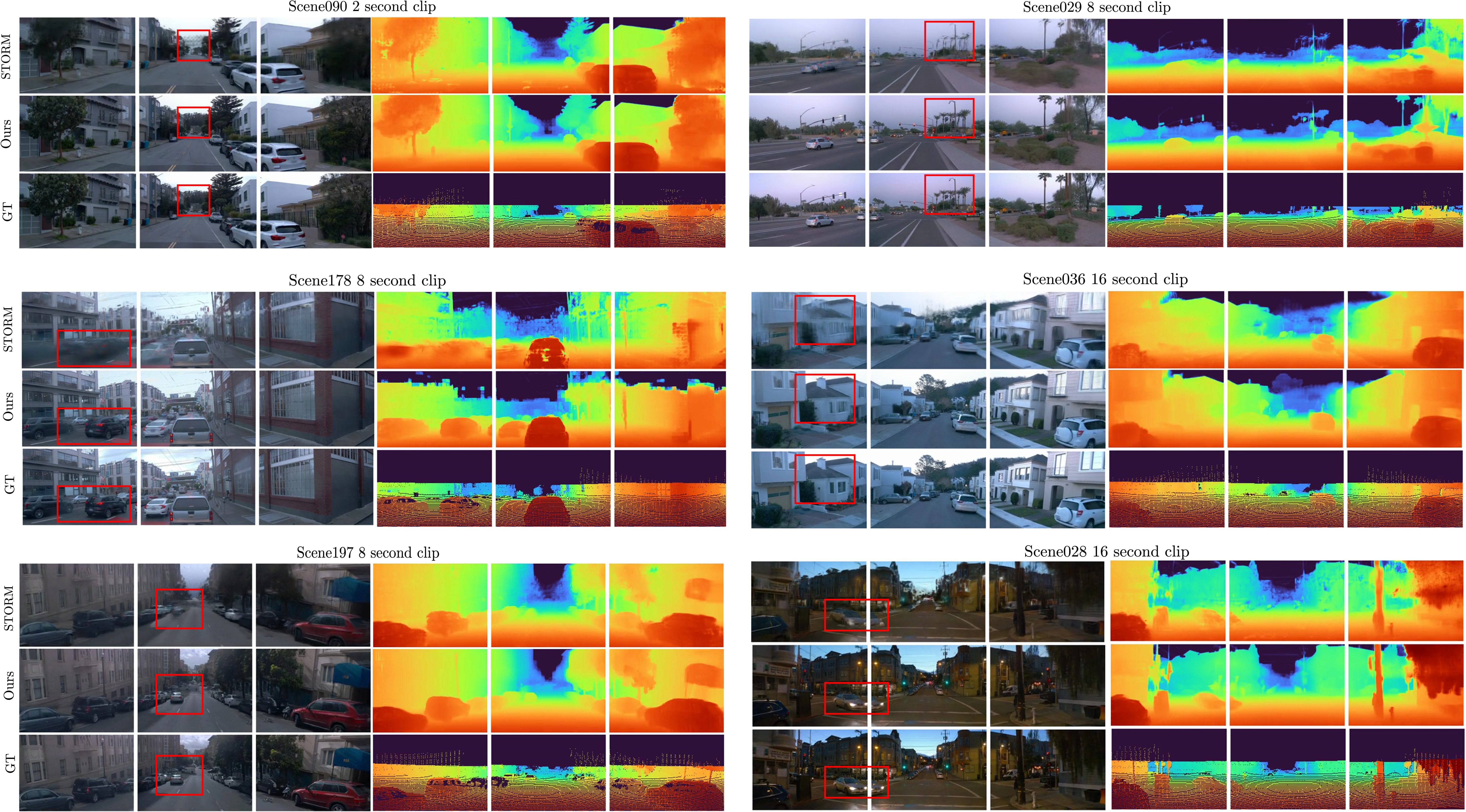}
    \caption{Additional Qualitative Results on Waymo Open Dataset.}
    \label{fig:qual_full}
\end{figure*}

We provide additional qualitative comparisons with STORM~\cite{storm} in Figure~\ref{fig:qual_full}. Our method demonstrates better visual quality and geometric accuracy in challenging scenarios, such as thin structures (e.g., poles and signboards) and moving objects (e.g., cars and pedestrians). Notably, STORM exhibits noticeable artifacts in distant regions and struggles to maintain consistent geometry for dynamic objects across extended temporal spans. In contrast, our recurrent refinement approach produces sharper details and more temporally coherent reconstructions. More importantly, our method scales better to longer sequences with complex dynamics, thanks to the recurrent scene update mechanism and hybrid dynamic modeling approach. 

\section{Recurrent Scene Update}
Algorithm~\ref{alg:ufo} provides a detailed specification of our recurrent scene update mechanism described in Section~\ref{sec:method} of the main paper. At each timestep $t$, the algorithm processes a new observation by: (1) selecting visible scene tokens from the previous state $\mathcal{S}_{t-1}$ based on camera frustum and proximity, (2) transforming these tokens to the local camera-centric coordinate system to stabilize learning, (3) applying the transformer update $\mathcal{T}_\text{update}$ to refine existing tokens and generate new ones, and (4) merging the refined and new tokens back into the global scene representation. The visibility filtering mechanism ensures computational efficiency by limiting attention to the most relevant $K$ tokens at each step, enabling near-linear scaling with sequence length. This selective updating strategy allows the model to iteratively refine previously observed geometry as new viewpoints become available, distinguishing our approach from standard feed-forward methods that lack refinement capabilities. Auxiliary tokens for sky modeling and affine color correction are maintained and updated throughout the sequence. After processing all timesteps, the final scene representation $\mathcal{S}_T$ is decoded into 3D Gaussians for rendering.

\begin{figure}[b]
    \centering
        \begin{algorithm}[H]
            \small
            \caption{Recurrent Scene Update for \method}
            \label{alg:ufo}
            \begin{algorithmic}[1]
                \Require Images $\{I_t\}_{t=1}^T$, poses $\{P_t\}_{t=1}^T$, intrinsics $\{\mathbf{K}_t\}_{t=1}^T$, tracked boxes $\{B_t\}_{t=1}^T$, visible-token budget $K$
                \State $\mathcal{S}_0 \gets \emptyset$ \Comment{Scene-token memory}
                \State $\theta_\text{sky}, \theta_\text{affine} \gets \text{InitAuxTokens}()$ \Comment{Shared auxiliary tokens}
                \For{$t = 1$ to $T$}
                    \State $R_t \gets \text{Pl\"uckerRays}(I_t, P_t, \mathbf{K}_t)$ \Comment{Per-pixel ray dirs from pose \& intrinsics}
                    \State $X_t \gets \text{ImageTokens}(I_t, R_t, t)$ \Comment{RGB + Pl\"ucker rays + time/pos. enc.}
                    \State $\mathcal{V}_{t-1} \gets \text{Visible}(\mathcal{S}_{t-1}, P_t, K)$ \Comment{Frustum filter + nearest $K$ tokens}
                    \State $\mathcal{V}_{t-1} \gets \text{ToLocal}(\mathcal{V}_{t-1}, P_t)$ \Comment{Camera-centric coordinates}
                    % \State $B_t^\star \gets \text{BBoxTokens}(B_t, t)$
                    \State $\mathcal{V}'_{t-1}, \Delta\mathcal{S}_t, \theta_\text{sky}, \theta_\text{affine} \gets \mathcal{T}_\text{update}(X_t, \mathcal{V}_{t-1}, B_t, \theta_\text{sky}, \theta_\text{affine})$
                    \State $\mathcal{S}_t \gets (\mathcal{S}_{t-1} \setminus \mathcal{V}_{t-1}) \cup \mathcal{V}'_{t-1} \cup \Delta\mathcal{S}_t$
                \EndFor
                \State $\mathcal{G}_T \gets \text{GaussianDecoder}(\mathcal{S}_T, \theta_\text{sky}, \theta_\text{affine})$
            \end{algorithmic}
        \end{algorithm}
\end{figure}

\end{document}